# Information Specialization and Constrained Synthesis in Multi-Agent LLM Forecasting: A Prospective Live-Study of the 2026 FIFA World Cup

Julian Varghese[1*] Lucas Bickmann[1] Sarah Sandmann[1]

[1]Institute of Medical Informatics, Otto-von-Guericke University Magdeburg, Germany.
Correspondence to: Julian Varghese <julian.varghese@med.ovgu.de>

**Abstract:** Large language models are being organized into multi-agent systems with specialized roles, but whether such specialization produces distinct forecasts and whether subsequent synthesis improves utility remains unclear. In this study, we carried out a live, prospective evaluation over the final 56 matches of the information-dense 2026 FIFA World Cup, keeping a frontier foundation model constant while assigning two primary forecasting agents contrasting specialist roles: a quantitative specialist focusing on structured performance statistics and a news specialist focusing on current injuries, tactics and information from press conferences. Their forecasts were then reviewed by a separate critic before being combined by a meta-agent, resulting in a sequential four-agent model. Forecasts from the betting market served as an external benchmark. The news specialist obtained the highest mean probability-weighted Top-3 utility and matched the betting market in Top-3 exact-score hits. Nevertheless, the two specialist forecasters agreed on at least two of the three scorelines in 50 out of 56 matches, and the meta-agent never generated more than one scoreline outside the specialists' forecast set. These findings show that rapidly changing, unstructured information can provide a valuable forecasting signal alongside structured statistics, whereas adding critic and meta-agent stages does not necessarily create complementary information or improve on the strongest specialist.

## Introduction

Large language models (LLMs) are becoming an integral part of multi-agent systems that distribute tasks across specialized agents and combine their outputs by means of interaction and critique[1–3]. However, multi-agent collaboration does not always turn out to be better than single-agent methods, and the advantages it offers will vary according to the nature of the task and the design of the system[4–7]. Hence, the key questions are 1) whether information-specialized roles within a fixed foundation model produce distinct forecasts, 2) how sequential critique and synthesis transform the forecast space generated by those roles, and 3) whether these differences translate into measurable differences in predictive performance. To look at this question prospectively, we make use of the recent 2026 FIFA World Cup, a highly prominent global event and therefore a unique real-world setting for forecasting. Predicting the results of the matches in this event offers a highly challenging environment for assessing this question, since the information to be processed is diverse, spread out over a large number of sources and is constantly changing. Structured sources offer recent performance data, expected-goal statistics, and team characteristics, while massively available unstructured news sources give information on injuries, line-ups, tactical decisions, recovery and press conferences. Moreover, the 2026 World Cup created a very information-dense environment of unstructured data - by the quarter-final stage, FIFA had reported about 42 million online and social-media stories and posts related to the tournament, more than 5,000 accredited media representatives and 282 official press conferences[8]. At the same time, official bookmaker market odds can be converted into market-based probability forecasts, providing an external reference against which the performance of the specialist and synthesized forecasts can be evaluated[9].

Recent studies have therefore used the 2026 World Cup for prospective LLM evaluation. Ding et al. evaluated four web-enabled frontier LLMs across all 104 matches and compared their win–draw–loss probabilities with betting-market forecasts, resulting in none of the LLMs surpassing the betting market in Brier score[10]. Zhaokai Wang et al. introduced

WorldCupArena, evaluating 13 language-model and deep-research systems on match results, exact scores, and additional football prediction tasks[11]. The best system showed only small improvements over betting-market and human baselines in match-result and exact-score accuracy, and they found no reliable improvement from adding web search. Zhenran Wang et al. separately evaluated six frontier LLMs across all 104 matches and found that their match-outcome performance was approximately on par with backing the bookmaker favorite, majority-vote aggregation did not improve performance[12]. While these studies primarily evaluate individual models or aggregate parallel forecasts, they do not isolate different information-specialized roles within a fixed foundation model—for example, a quantitative agent focused on structured performance data and a news agent focused on contemporaneous unstructured information—and then examine how those specialist forecasts are transformed by subsequent critique and synthesis. This design enables the effects of information specialization to be distinguished from those of downstream synthesis. Moreover, less is known about exact-score forecasting, which is more granular than win–draw–loss prediction and empirically more difficult[11]. Unlike a single exact-score prediction, a ranked Top-k forecast retains information about several plausible outcomes and their relative probabilities, which is why Top-k prediction is well established in multiclass machine learning as a complement to strict Top-1 evaluation[13]. To retain information about both forecast probability and utility, ranked exact-score predictions can additionally be evaluated using a probability-weighted outcome utility.

We developed a sequential multi-agent system with separation by design according to information type while holding the underlying foundation model fixed. This controls for differences in underlying model capability and reflects a practical use pattern in which the same LLM is prompted sequentially with different information-focused roles and synthesis instructions. A quantitative specialist (Agent A) focused primarily on structured performance information, whereas a news specialist (Agent B) focused on contemporaneous unstructured information; their forecasts were then reviewed by a critic (Agent C) and adjudicated by a meta-agent (Agent E). Retaining the outputs of each stage prospectively enabled direct within-match comparison of the specialist forecasts and assessment of how sequential synthesis modified the forecast space generated upstream and whether it improved predictive performance. Each forecasting agent produced three ranked exact-score predictions with associated probabilities before match outcomes were known, and all forecasts were recorded in a public, timestamped, version-controlled GitHub ledger. The agents were instructed not to use betting odds, implied probabilities, or bookmaker predictions. Bookmaker-derived forecasts were collected separately as an external market benchmark. To evaluate ranked predictions jointly with their probabilities, we used a graded scoreline utility derived from the official UEFA Predictor rules, which assigns partial credit for categorical agreement with the observed outcome rather than exact-score correctness alone.

The novelty of our multi-agent approach lies in a live, prospective study that separates quantitative and contemporaneous news roles within a fixed foundation model during the exceptionally information-rich setting of the FIFA World Cup. By retaining the individual specialist and meta-agent forecasts for every match, the design enables direct comparison of how the quantitative and news roles differ and whether subsequent critique and synthesis preserve, change, or expand the forecast space.

# Results

## Multi-agent forecasting framework

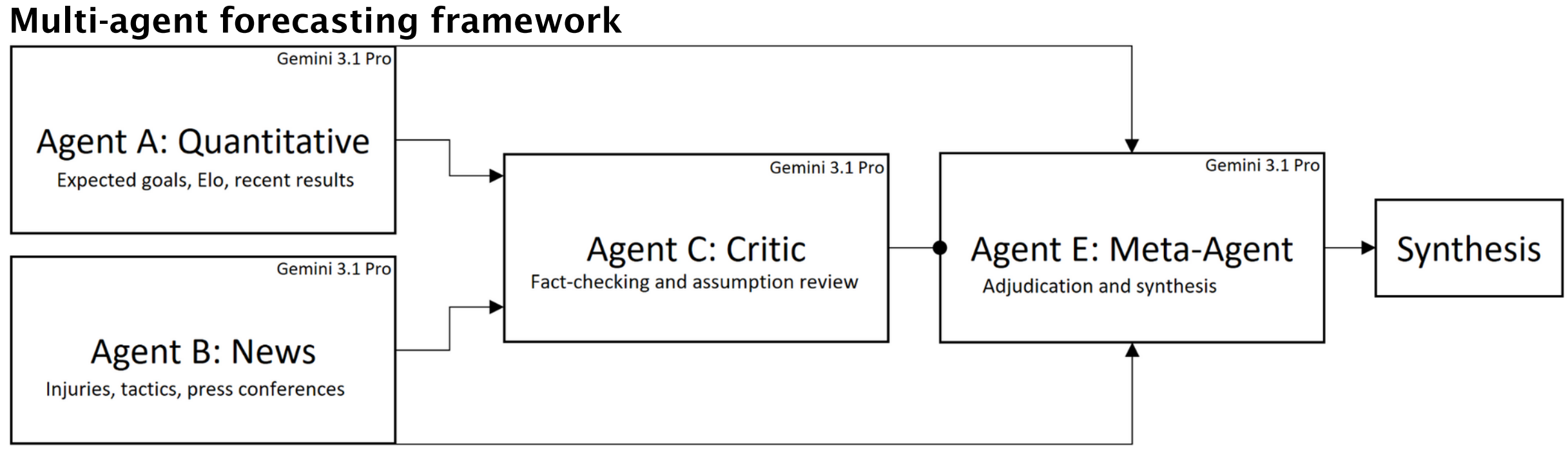

**Figure 1:** Multi-agent architecture. For each match, the quantitative and news agents independently generated ranked Top-3 exact-score predictions with associated probabilities and reasoning explanations. The critic agent reviewed these outputs and produced an assessment, after which the specialist predictions and the critic's evaluation were passed to the meta-agent to generate its synthesis.

The sequential architecture is illustrated as Figure 1. The starting point of forecasts was chosen to ensure that agent forecasts could draw on a reasonable amount of current tournament-specific evidence for each team, including two completed matches per team as well as contemporaneous news and contextual information. Therefore, forecasting and the resulting evaluation comprised the last 56 matches of all 104 matches. For each included match, the quantitative (agent A) and news-specialized agent (agent B) generated probabilistic exact-score Top-3 predictions for each match, followed by a critical review (agent C) providing only free-text review of the first two agents without Top-3 predictions and the final meta-analytical agent (agent D) to conclude its exact-score Top-3 prediction based on the output of all three preceding agents. Contemporaneous exact-score betting-market estimates were collected separately as an external benchmark from oddsportal.com[14]. Both the agents outputs and the betting-market data were collected 19-20h before the corresponding match and then manually checked to correct minor formatting errors. All evaluated forecasts were timestamped before the corresponding match outcomes and pushed to the version-controlled prediction ledger[15].

**Evaluation**
Forecasting performance was evaluated using four complementary measures that capture both exact-score accuracy and the quality of near-miss predictions (Table 1). 1) The primary measure was the probability-weighted Top-3 utility (PwU), which combines the three highest-ranked exact-score forecasts with their assigned probabilities and the official UEFA utility score for a predicted outcome. 2) Top-3 exact hit rate provides a stricter probability-independent measure and records whether the observed exact score appeared anywhere among the three highest-ranked predictions. 3) Top-1 utility evaluates only the single most likely scoreline using the UEFA scoring, whereas 4) Top-1 exact accuracy records whether this single prediction exactly matched the observed score. In addition to the three agent-based forecasters and the betting-market benchmark, we included an unconditional football prior—a fixed baseline derived from general football scoreline frequencies and therefore independent of the teams or match-specific information.

Table 1: Evaluation of performance measures for all 56 matches. PwU: Probability-weighted utility (Primary Measure).

| Forecaster | Top-3 PwU mean (SD) | Top-3 exact hit, n/N (%) | Top-1 utility, mean (SD) | Top-1 exact accuracy, n/N (%) |
|---|---|---|---|---|
| Quantitative agent (A) | 1.009 (0.537) | 20/56 (35.7%) | 3.107 (1.836) | 7/56 (12.5%) |
| News agent (B) | **1.066** (0.588) | **23/56** (41.1%) | 2.714 (2.025) | 7/56 (12.5%) |
| Meta-agent (E) | 1.030 (0.580) | 18/56 (32.1%) | 2.732 (1.892) | 5/56 (8.9%) |
| Betting market | 1.033 (0.559) | **23/56** (41.1%) | **3.214** (2.078) | **11/56** (19.6%) |
| Unconditional football prior | 0.607 (0.196) | 14/56 (25.0%) | 1.768 (2.123) | 8/56 (14.3%) |

The news agent and the betting market achieved the highest mean Top-3 PWU (1.066, 1.033) Top-3 exact-score hits (23/56; 41.1%), whereas the betting market performed best on both Top-1 metrics. The meta-agent did not exceed the strongest specialist on any summary measure. Total Top-3 probability mass was similar across the LLM forecasters (A, 0.382 ± 0.060; B, 0.386 ± 0.061; E, 0.393 ± 0.082); the corrected betting market mass was somewhat lower (0.359 ± 0.052), which should be considered when interpreting the small PwU difference between the news agent and market.

**Cumulative Utility**

Cumulative utility trajectories remained closely aligned throughout the tournament (Figure 2). The betting market showed an early advantage during the first few matches, but this advantage disappeared as the tournament progressed. The meta-agent subsequently showed a temporary mid-tournament lead before converging with the other forecasters, whereas the news agent (Agent B) moved ahead late in the tournament and finished with the highest cumulative utility. In contrast, the quantitative agent (Agent A) performed less well than the other forecasters, with the gap becoming more pronounced toward the end of the evaluation period. The unconditional football prior separated early from the informed forecasters, with the performance gap progressively widening over the course of the tournament.

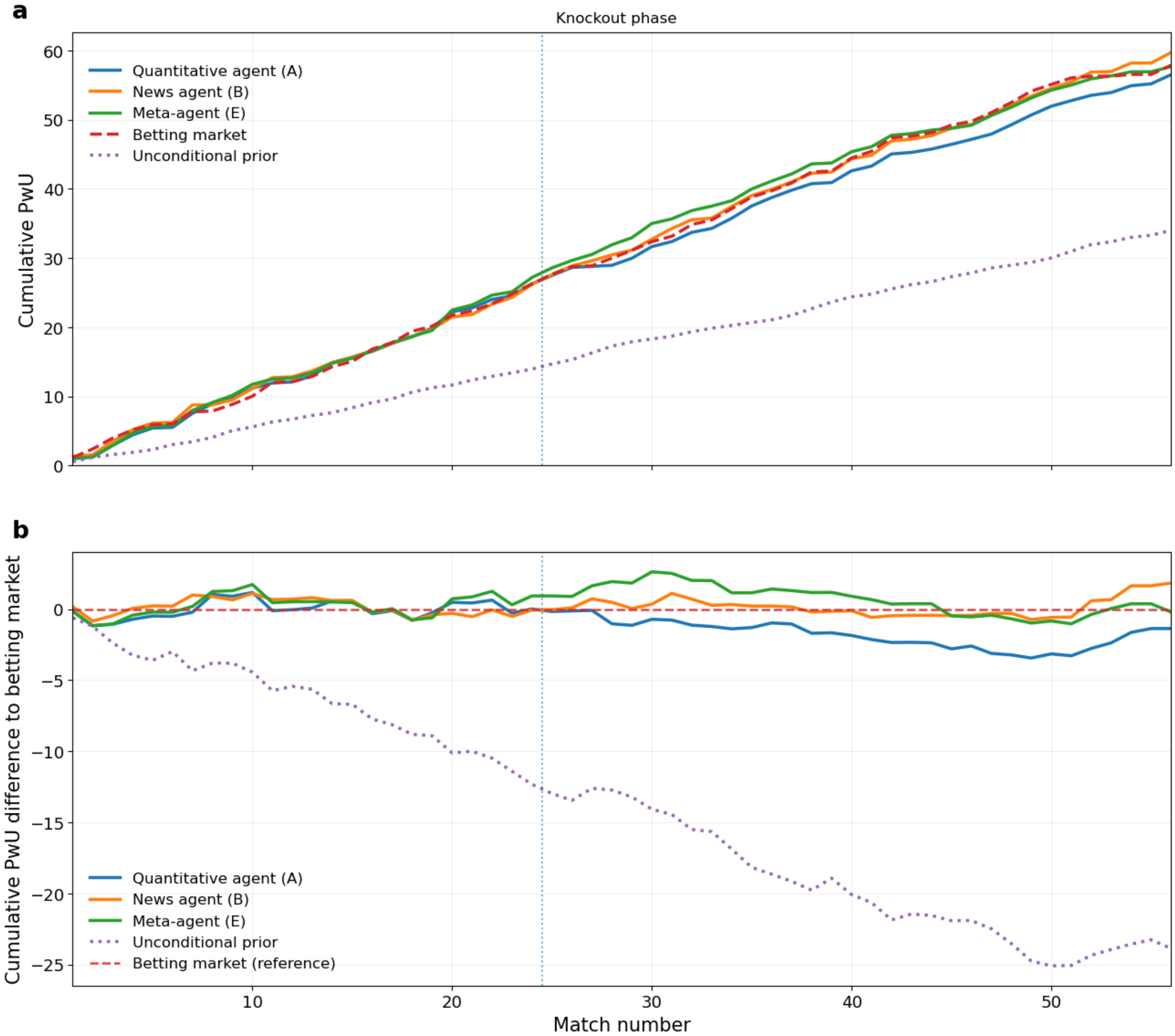

**Figure 2:** Cumulation of Top3-probability-weighted utility (PwU) progress during the tournament among competitors (a) and differences to the betting market specifically (b).

**Prediction overlaps between competitors**

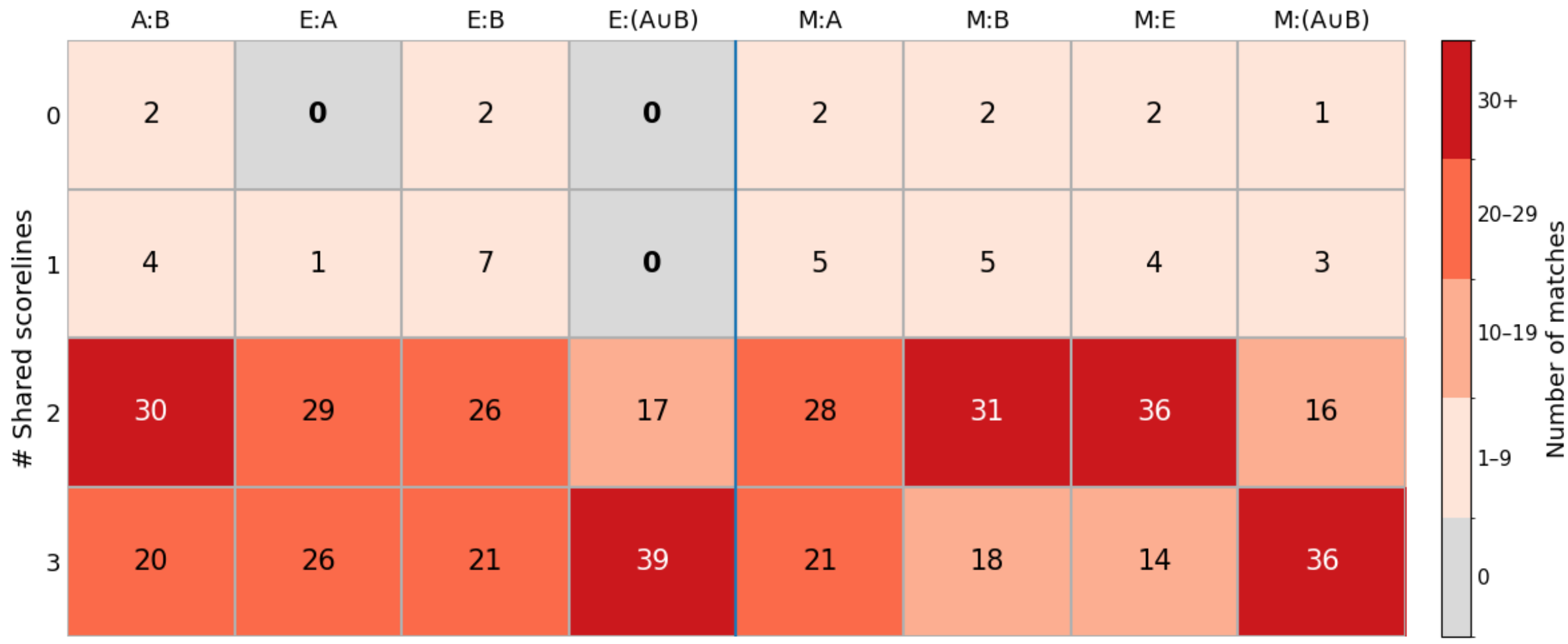

**Figure 3:** Exact Top-3 scoreline overlaps in all 56 matches. For instance, the first column (A:B) compares the exact match Top-3- scorelines of A and B. In 2 matches no shared scorelines were observed, meaning they predicted 3 different scorelines. In 4 matches they shared one exact scoreline, in 30 matches 2, in 20 they all shared three scorelines. A: Quantitative agent, B: News agent, E: Meta-agent, M: Betting market.

The quantitative and news agents shared at least two of their three exact-score predictions in 50/56 matches (89.3%). The meta-agent remained particularly close to the combined specialist forecast set: at least 2 of its three scorelines were contained in union of the specialists' outputs in all 56 matches, including all three in 39 out 56 matches (69.6%) (Figure 3, column 4). The meta-agent never shared fewer than two of its three predicted scorelines with the combined specialists outputs. In other words, it never never generated more than one scoreline outside the combined specialist forecast set, whereas the betting market did so in 4/56 matches.
A semantically substantial outcome-level deviation (e.g. a win instead of a loss or draw) from the specialist union was technically possible in 41 matches (73.2%), as in these matches not all outcomes were covered (win, loss, draw). The meta-agent made such a deviation in only 6 of these 41 possible matches. For these 6, the meta agents utility (PwU) was still in between both specialists in three cases, in one case it was better, two cases worse.

# Discussion

The informed LLM forecasts achieved probability-weighted Top-3 utilities in the same range as the betting-market reference and substantially above the unconditional football prior. Among the LLM conditions, the news specialist achieved the highest mean PwU (1.066) and matched the betting market in Top-3 exact-score hits (23/56; 41.1%), whereas the market remained strongest on the Top-1 measures. The higher PwU of the news specialist relative to the other LLM agents was not simply attributable to greater Top-3 probability mass, which was similar across A, B and E. The difference between the news and quantitative specialists may reflect the predictive value of rapidly changing, match-specific information that is only incompletely represented by historical performance statistics. Previous work has demonstrated that incorporating contextual information from football journalism alongside statistical match data can improve match-outcome prediction relative to traditional statistical approaches[16]. Moreover, this differs from the broader question of whether web search itself improves forecasting: WorldCupArena found no reliable forecasting advantage from adding self-directed websearch[11]. Our findings instead raise the possibility that the selection and specialization of retrieved information may matter more than access to web search in general.

The utility performances changed over the course of the tournament. The betting market accumulated the highest PwU early in the evaluation, the meta-agent temporarily moved ahead during the middle portion, and the news specialist finished with the highest cumulative PWU. The meta-agent's temporary advantage coincided with the transition from the final group-stage round to the knockout phase [17] . Although exploratory, this pattern raises the possibility that synthesis was transiently useful during a period in which newly accumulated tournament information had to be reconciled with a changing competitive context.

The internal forecasts revealed substantial constraints on how the downstream synthesis stage transformed the specialist outputs. The quantitative and news agents already shared at least two exact scorelines in 50/56 matches, and the meta-agent retained at least two scorelines from their combined forecast set in every match, including all three in 39/56. In none of the 56 matches did the meta-agent generate more than one scoreline outside the combined specialist forecast set. By comparison, the independent betting-market reference produced at least two scorelines outside the specialist union in 4 of 56 matches. This indicates that the meta-agent rarely generated genuinely different scorelines. Moreover, although a substantial outcome-level deviation from the specialist union was possible in 41/56 matches, the meta-agent made such a departure in only 6/41 opportunities. Together, these observations indicate that in this architecture, downstream synthesis predominantly selected or locally modified hypotheses already represented by the specialists rather than substantially expanding the forecast space. This limited forecast diversity may also have reduced the potential benefit of synthesis: forecast combination is most beneficial when component forecasts provide accurate and sufficiently diverse information, whereas correlated or weaker forecasts can dilute the quality of the combined prediction[18]. This may be particularly relevant here because all agents were role-specialized instances of the same underlying model. Consistent with this interpretation, Zhenran Wang et al. found no benefit from majority-vote aggregation across frontier LLM forecasts[12], and Kim et al. more generally showed that multi-agent coordination can either improve or deteriorate performance depending on the model's capabilities, the task, and the architecture, with error amplification being one of the identified failure modes[5]. The meta-agent may therefore have combined highly overlapping specialist forecasts without adding sufficiently distinct information to consistently improve on the stronger specialist.

In conclusion, our findings suggest that information specialization and downstream synthesis represent distinct design problems in multi-agent LLM forecasting. Information-specialized roles produced competitively performing forecasts within a fixed foundation model, with the news specialist achieving the highest mean PwU. However, subsequent critique and synthesis did not consistently improve on the strongest specialist and rarely expanded the forecast space generated upstream, even when outcome-level departures were available. These findings suggest that adding sequential agent stages does not inherently create complementary predictive information; their value may depend on whether role specialization generates sufficiently distinct signals and whether the synthesis mechanism can identify when and how to depart from them. Further prospective evaluations across models, architectures and forecasting domains are needed to determine how broadly these patterns generalize.

# Methods

## Prospective Study Design

A prospective forecasting study was carried out during the 2026 FIFA World Cup. The forecasting process started on the third day of the group stage and went on until the final, covering a total of 56 matches. By beginning the forecasts from the third group stage matchday, it was ensured that each team involved had already played two matches in the tournament before the forecasting began, so that all teams had a common minimum amount of tournament-specific information regarding their recent performance, current form, tactical tendencies, and up-to-date team news. This approach also meant that it was not necessary to compare forecasts that had been made based on very different levels of tournament-specific information during the first matchdays. For all the agents, the predictions referred only to regular time including stoppage time and not to extra time or penalty shoot-outs, in order to be in line with the exact-score betting market benchmark, which itself only included regular-time results. The agents' outputs were given in a predefined structured JSON format consisting of a short explanation of the reasoning and three scorelines ranked in order with their respective probabilities. The information restrictions were applied as explicit prompt-level guardrails and not as technical domain filters. For the bookmakers' data, the three highest-ranked exact-score odds for each match were obtained from oddsportal.com. The agent outputs, together with the reasoning and the predicted scorelines, were collected in sequence about 20 hours before the official start of the match and then pushed to the public ledger (predictions_ledger.jsonl) after a brief manual check. This check consisted only in correcting any syntactic or formatting errors that were necessary for proper structured parsing during the final evaluation. The predicted scorelines, their rankings, probabilities, and the main reasons for the predictions were left unchanged. All entries into the public ledger were timestamped in UTC and took place before the scheduled kickoff of the relevant match.

## Forecasting Framework

The full agent prompts and model specifications are included in the code in the files Main-Analysis and agent_module respectively[15]. All agents within the multi-agent framework were implemented in Python 3.11.11 by making separate API calls to the same underlying LLM model Gemini 3.1 Pro Preview, via the Google GenAI package version 1.41.0. This model was chosen since it supports complex reasoning including calculations, web-based information retrieval, and multi-step synthesis. Google Search grounding was activated for each agent call in order to allow access to up-to-date public information. A generation temperature of 0.2 was used for all the calls, and no specific thinking level parameter was set.

The quantitative specialist, Agent A, was told to base its evaluation mainly on structured football data, such as the teams' most recent match results and short-term expected-goal trends, their attacking and defensive strengths, head-to-head records, squad market values, Elo ratings, the teams' rest and travel conditions. Web searches were specifically limited to statistical websites like FBref, Transfermarkt and the official FIFA statistics, and it was forbidden to look for predictions, betting odds or betting advice. The agent estimated the team-specific expected-goal parameters ($\lambda 1$, $\lambda 2$) by assuming an independent-Poisson model – a method used for modelling the number of goals in low-scoring football matches – and then provided three ranked exact-score forecasts together with the probabilities calculated by the model. Agent B, the news specialist, was intended to pick up on quickly changing contextual details and was asked to search mainly for information from the previous 48 hours, covering team news, press conferences, injuries, player availability, current form, tactical match-ups, set pieces, environmental conditions and squad depth. It was clearly instructed not to consider bookmaker information and to exclude market sentiment from the material it retrieved. Instead of building its own quantitative historical model from scratch, it converted the contextual evidence it had gathered into adjustments of the expected-goal parameters and produced its own ranked Top-3 exact-score forecast. Agent C, the critic, was given the full outputs from both specialist agents and examined the evidence and the assumptions on which their expected-goal estimates were based. It did not prepare an independent forecast. Its access to the web was restricted to checking the specific claims made by the specialists—for instance, verifying an injury or a statistical statement—and it was explicitly prohibited from searching for general match predictions or from using betting-market information to check either of the forecasts. Agent D, the meta-agent, was given both the specialist forecasts and the critic's evaluation and produced the final ranked probabilistic prediction. The prompt included a clear anti-averaging rule: the meta-agent was instructed not to simply average the specialists' probabilities or their expected-goal estimates, but rather to assess the evidence, accept or reject the individual critiques and to recalibrate the expected-goal parameters where appropriate. At this stage, web searches were limited to resolving specific factual disputes, while betting odds, consensus predictions and market sentiment were explicitly excluded from the calibration process.

## Evaluation

The analysis scripts for generating all result table values and images are provided as part of the code in the file evaluation_ledger[15].

### Forecasting Performance Measures

We evaluated forecasting performance using four complementary measures designed to capture different aspects of prediction quality in football match outcomes. Together, they balance strict exact-score correctness with partial credit for near-correct predictions and distinguish the quality of the single most likely forecast from the broader information contained in the ranked Top-3 predictions.

Combination reduces dependence on any single scoring rule and provides a more robust assessment of both forecast accuracy and probabilistic ranking quality.

Top-3 Probability-weighted Utility (PwU): For each match m, the model's three highest-ranked scoreline predictions were evaluated jointly by weighting the UEFA utility of each scoreline by its corresponding model-generated probability:

$$\mathrm{PwU}_m = \sum_{i=1}^{3} p_{mi}\, U(s_{mi}, y_m).$$

$U(s_{mi}, y_m)$ is the UEFA utility[19] of predicted scoreline $s_{mi}$ relative to the observed score $y_m$. A correct match outcome is worth 3 points, with 1 additional point each for the correct first-team goal count, the correct second-team goal count, and the correct goal difference. An exact score therefore receives 6 points.
$p_{mi}$ denotes the forecast probability returned by the model as the i-th ranked scoreline $s_{mi}$ for match m. In order to account for potential probability inflation across forecasters, we additionally analyzed the probability mass assigned to each Top-3 prediction set.

When converting bookmaker odds into probabilities, a correction is required because quoted odds incorporate the bookmaker's overround and therefore do not correspond directly to unbiased event probabilities. We therefore converted decimal correct-score odds using p=1/(1.12 × odds), applying the reference booksum of 1.12, which corresponds to the scoreline overround reported for oddsportal.com by Reade et al. (2020)[20].

Top1-Utility: Only the single highest-ranked (=Top1) scoreline was evaluated using the UEFA-derived utility.

$$\mathrm{T1U}_m = U(s_{m1}, y_m).$$

Top-3 exact-score hit rate: This measure records whether the observed exact score appeared anywhere among the three highest-ranked predictions:

$$H_m^{(3)} = \mathbb{I}\{y_m \in \{s_{m1}, s_{m2}, s_{m3}\}\}$$

Top-1 exact-score accuracy: This strict measure records whether the highest-ranked predicted scoreline exactly matched the observed result:

$$H_m^{(1)} = \mathbb{I}\{s_{m1} = y_m\}$$

As an uninformed baseline for the prior, we used the historical World Cup scoring rate given by Chater et al. (2021)[21], which was α=2.5156 goals per match. On the assumption that the two teams were equal in strength, each was given a value of λ=α/2=1.2578. The probabilities for exact scores were calculated from the joint distribution of two independent Poisson-distributed goal totals, and the three scorelines with the highest probabilities were selected as the Top-3 forecast. The same distribution was used for each match without incorporating any team-specific information.

**Forecast-space overlap**

In order to measure the similarity between the forecast outputs including the bookmaker, the Top-3 scorelines were regarded as unordered sets. We further examined whether the meta-agent departed from the outcome-level space represented by the two specialists (e.g. win instead of loss or draw), which is numerically determined by similarity <2 using the UEFA-derived scoring. We additionally determined, for each match, whether such a deviation by the meta-agent was available among the distinct scorelines observed across the prospective forecasting ledger and if the meta-agent deviated how did it affect utility.

**Statistical Analysis**

All analyses were conducted in Python 3.11.11 and are included in the file evaluation_ledger. The unit of evaluation was the match (n=56); all forecasters were evaluated on the same prospectively defined match cohort and comparisons were therefore paired. Performance metrics are summarized descriptively as mean (SD) or n/N (%), and central architecture comparisons are additionally expressed as within-match differences in PwU. Because the evaluated matches constituted the complete prespecified tournament cohort rather than an independent random sample, inferential tests were not used to establish differences between forecasters. Exploratory forecast-space and outcome-level deviation analyses were descriptive. As a sensitivity analysis for the probability weighting of PwU, total Top-3 probability mass was summarized for each forecaster.

## Acknowledgement

GPT-5.6 Sol (OpenAI) was used for AI-assisted copy editing of author-generated text to improve readability and language. The authors reviewed and approved the final manuscript. We acknowledge support by the Open Access Publication fund of the medical faculty of the Otto-von-Guericke-University Magdeburg.

## References

1. Du Y, Li S, Torralba A, Tenenbaum JB, Mordatch I. Improving Factuality and Reasoning in Language Models through Multiagent Debate. International Conference on Machine Learning. 2024;235:11733–63. doi:10.48550/arXiv.2305.14325

2. He J, Treude C, Lo D. LLM-Based Multi-Agent Systems for Software Engineering: Literature Review, Vision, and the Road Ahead. ACM Trans Softw Eng Methodol. 2025 Jun 30;34(5):1–30. doi:10.1145/3712003

3. Li X, Wang S, Zeng S, Wu Y, Yang Y. A survey on LLM-based multi-agent systems: workflow, infrastructure, and challenges. Vicinagearth. 2024 Oct 8;1(1):9. doi:10.1007/s44336-024-00009-2

4. Wang Q, Wang Z, Su Y, Tong H, Song Y. Rethinking the bounds of llm reasoning: Are multi-agent discussions the key? In: Proceedings of the 62nd Annual Meeting of the Association for Computational Linguistics (Volume 1: Long Papers) [Internet]. 2024 [cited 2026 Aug 26]. p. 6106–31. Available from: https://aclanthology.org/2024.acl-long.331/

5. Kim Y, Gu K, Park C, Park C, Schmidgall S, Heydari AA, et al. Capable language models can outgrow the benefits of collaboration. Nat Mach Intell. 2026 Jul;8(7):1157–72. doi:10.1038/s42256-026-01268-y

6. Gómez Álvarez S, Mozo Quesada A, Navarro T, Gálvez Rojas S, López Valverde F. Multi-Agent debate system based on large language models: structured deliberation and validation in satellite communications. J Intell Inf Syst. 2026 Aug 13. doi:10.1007/s10844-026-01086-z

7. Klang E, Omar M, Raut G, Agbareia R, Timsina P, Freeman R, et al. Orchestrated multi agents sustain accuracy under clinical-scale workloads compared to a single agent. npj Health Syst. 2026 Mar 9;3(1):23. doi:10.1038/s44401-026-00077-0

8. From packed stadiums to record digital reach: FIFA World Cup 2026™ numbers tell story of unprecedented scale as last eight confirmed [Internet]. [cited 2026 Aug 26]. Available from: https://inside.fifa.com/organisation/media-releases/origin1904-p.cxm.fifa.com/packed-stadiums-record-digital-reach-world-cup-2026-numbers-unprecedented-scale

9. Štrumbelj E. On determining probability forecasts from betting odds. International Journal of Forecasting. 2014 Oct 1;30(4):934–43. doi:10.1016/j.ijforecast.2014.02.008

10. Ding J, Guo C, Xu J. FIFA World Cup 2026 as a Contamination-Free Benchmark for LLM Forecasting Agents: Four Models, a Bookmaker, and 104 Matches [Internet]. arXiv; 2026

[cited 2026 Aug 26]. Available from: http://arxiv.org/abs/2607.17765 doi:10.48550/arXiv.2607.17765

11. Wang Z, Gui T, Rao J, Di S, Tang Y, Liang D. WorldCupArena: Fine-Grained Evaluation of Language Models and Deep-Research Agents on Football Forecasting [Internet]. arXiv; 2026 [cited 2026 Aug 26]. Available from: http://arxiv.org/abs/2607.18084 doi:10.48550/arXiv.2607.18084

12. Wang Z, Bian Z, Li J, Qi Z. WorldCup Arena: Prospective, Leakage-Free Evaluation of Frontier LLMs on a Live Tournament [Internet]. arXiv; 2026 [cited 2026 Aug 26]. Available from: http://arxiv.org/abs/2608.04008 doi:10.48550/arXiv.2608.04008

13. Berrada L, Zisserman A, Kumar MP. Smooth Loss Functions for Deep Top-k Classification [Internet]. arXiv; 2018 [cited 2026 Aug 26]. Available from: http://arxiv.org/abs/1802.07595 doi:10.48550/arXiv.1802.07595

14. s.r.o L. Oddsportal.com [Internet]. [cited 2026 Aug 26]. Odds Comparison, Sports Betting Odds | OddsPortal. Available from: https://www.oddsportal.com/

15. VargheseLab. GitHub [Internet]. [cited 2026 Aug 26]. GitHub - VargheseLab/Fifa2026-Pred at update-files-branch. Available from: https://github.com/VargheseLab/Fifa2026-Pred/tree/update-files-branch

16. Beal R, Middleton SE, Norman TJ, Ramchurn SD. Combining machine learning and human experts to predict match outcomes in football: A baseline model. In: Proceedings of the AAAI Conference on Artificial Intelligence [Internet]. 2021 [cited 2026 Aug 26]. p. 15447–51. Available from: https://ojs.aaai.org/index.php/AAAI/article/view/17815

17. World Cup 2026 Knockout Stages: All the results up to the final [Internet]. [cited 2026 Aug 26]. Available from: https://www.fifa.com/en/tournaments/mens/worldcup/canadamexicousa2026/articles/knockout-stage-match-schedule-bracket

18. Lichtendahl Jr KC, Winkler RL. Why do some combinations perform better than others? International Journal of Forecasting. 2020;36(1):142–9.

19. UEFA.com. UEFA.com [Internet]. 2023 [cited 2026 Aug 26]. UEFA Champions League Predictor rules 2023/24 | UEFA Champions League 2023/24. Available from: https://www.uefa.com/uefachampionsleague/news/0278-15df8707b234-9cebd98d051d-1000--uefa-champions-league-predictor-rules-2023-24/

20. Reade J, Singleton C, Vaughan Williams L. Betting markets for English Premier League results and scorelines: evaluating a forecasting model. Economic Issues. 2020;25(1):87–106.

21. Chater M, Arrondel L, Gayant JP, Laslier JF. Fixing match-fixing: Optimal schedules to promote competitiveness. European Journal of Operational Research. 2021 Oct 16;294(2):673–83. doi:10.1016/j.ejor.2021.02.006